\pdfoutput=1

\documentclass[11pt]{article}

\usepackage[preprint]{acl}

\usepackage{times}
\usepackage{latexsym}

\usepackage[T1]{fontenc}

\usepackage[utf8]{inputenc}

\usepackage{microtype}

\usepackage{inconsolata}

\usepackage{graphicx}

\usepackage{tabularray}

\title{Inspecting the Representation Manifold of Differentially-Private Text}

\author{Stefan Arnold \\ 
Friedrich-Alexander-Universität Erlangen-Nürnberg \\
Lange Gasse 20, 90403 Nürnberg, Germany \\ 
\texttt{stefan.st.arnold@fau.de}}

\begin{document}
\maketitle
\begin{abstract}

Differential Privacy (DP) for text has recently taken the form of text paraphrasing using language models and temperature sampling to better balance privacy and utility. However, the geometric distortion of DP regarding the structure and complexity in the representation space remains unexplored. By estimating the intrinsic dimension of paraphrased text across varying privacy budgets, we find that word-level methods severely raise the representation manifold, while sentence-level methods produce paraphrases whose manifolds are topologically more consistent with human-written paraphrases. Among sentence-level methods, masked paraphrasing, compared to causal paraphrasing, demonstrates superior preservation of structural complexity, suggesting that autoregressive generation propagates distortions from unnatural word choices that cascade and inflate the representation space.



\end{abstract}

\section{Introduction}

\textit{Language Models} (LMs) \citep{chowdhery2023palm} are trained on extensive corpora of text containing sensitive information. Several studies demonstrated that sensitive information can be extracted from LMs \citep{song2019auditing, pan2020privacy, nasr2023scalable, carlini2023quantifying}, raising significant privacy concerns and prompting the integration of privacy mechanisms. 

To protect against unintended disclosure of information, \textit{Differential Privacy} (DP) \citep{dwork2006calibrating} has been tailored to raw text \citep{fernandes2019generalised, feyisetan2020privacy}. Through a randomized mechanism, 
DP formalizes privacy through a notion of indistinguishability, ensuring that texts remain statistically unaffected by the addition or removal of individual samples in the text corpus. 

While early randomized mechanisms exploit the distances between words in the embedding space \citep{mikolov2013efficient} to replace words with a noisy approximation of their nearest neighbor, grammatical constraints associated with word-level privatization \citep{mattern2022limits} has led to a shift towards paraphrasing text at sentence-level by leveraging LMs \citep{igamberdiev2023dp, utpala2023locally, meisenbacher2024dp}.

\paragraph{Contribution.} We inspect the representation geometry of text paraphrased under the privacy constraints of DP, accounting for different levels of privacy. \citet{ansuini2019intrinsic} discovered that high-dimensional signals reside on low-dimensional manifolds, a property that holds across neural representations \citep{tulchinskii2024intrinsic}. Building on \textit{Intrinsic Dimensionality} (ID), we estimate the ID of texts and interpret ID shifts as a proxy for distortions on their structure and complexity. Specifically, we compare differentially-private transformations operating on word-level and sentence-level. We find that word-level DP deviates the most from human-authored paraphrases, significantly altering the underlying representation space. Concerning sentence-level DP, we argue that bidirectional paraphrasing based on masked substitution mitigates cascading errors that arise in sequential generation.


\section{Background}
\label{sec:background}

We briefly provide the necessary foundations for differential privacy and intrinsic dimensionality.

\subsection{Differential Privacy}

\textit{Differential Privacy} (DP) is a notion of privacy introduced by \citet{dwork2006calibrating} under the term $\varepsilon$-indistinguishability. DP operates on the principle of adding noise calibrated to the sensitivity of adjacent datasets that differ by at most one record. The level of indistinguishability can be controlled by the privacy budget $\varepsilon \in (0,\infty]$, with declining privacy guarantees as $\varepsilon \rightarrow \infty$.

To mitigate the disclosure of authorship \citep{song2019auditing}, DP is applied to perturb raw text either at word level or sentence level through noise injected into embedding models \citep{mikolov2013efficient} and language models \citep{peters2018deep, radford2018improving}, respectively.

\paragraph{Word-level DP.} 

\citet{feyisetan2020privacy} introduced a randomized mechanism in which a text is perturbed at the word level by mapping each word to another word located within a radius derived from an embedding space and governed by the privacy budget $\varepsilon$. This randomized mechanism was termed \texttt{MADLIB}. By scaling the notion of indistinguishability by a distance, \texttt{MADLIB} satisfies the axioms of metric DP \citep{chatzikokolakis2013broadening}. Despite many refinements regarding the preservation of utility \citep{carvalho2021tem, xu2021utilitarian, yue2021differential} and privacy \citep{xu2020differentially, xu2021density}, \texttt{MADLIB} continues to suffers from syntactic errors \citep{mattern2022limits} and semantic drift \citep{arnold2023driving}.

\paragraph{Sentence-level DP.}

Given the shortcomings of \texttt{MADLIB} and its recent refinements \citep{yue2021differential, chen2023customized}, researchers conceptualized the privatization of text as paraphrasing by utilizing sequence-to-sequence models \citep{bo2021er, krishna2021adept, weggenmann2022dp, igamberdiev2023dp}. Unlike word-level mechanisms, which perturb text on a word-by-word basis, sentence-level mechanisms paraphrase entire sentences. A defining characteristic shared is the injection of noise into the encoder representations, and learning of the decoder to generate fluent paraphrases while obfuscating stylistic identifiers that could otherwise compromise privacy.


\citet{mattern2022limits} conjectured that temperature sampling in LMs can be interpreted as an instance of the exponential mechanism \citep{mcsherry2007mechanism}, where the scoring function corresponds to most probable word given a context. The probability of selecting a word follows the softmax distribution over the \textit{logits}, which represent the likelihood of each word occurring in a given context. Since DP requires the sensitivity to be bounded, these logits are clipped in range. 

Since paraphrasing is contingent upon the resemblance between the training text and the text subjected to privatization, \citet{utpala2023locally} leverage the generalization capabilities of large-scale pre-trained LMs to generate paraphrases via zero-shot prompting. \citet{meisenbacher2024dp} depart from autoregressive generation and instead adopted the idea of temperature sampling to masked LMs. Unlike causal LMs, which sample text sequentially, this approach masks words and predicts its substitution bidirectionally from context.

\subsection{Intrinsic Dimensionality}

Grounded on the manifold hypothesis \citep{fefferman2016testing}, the concept of intrinsic dimensionality characterizes the number of degrees of freedom for data in a representation space. Unlike extrinsic dimensionality, which corresponds to the overall dimensionality of the representation space, the intrinsic dimension (ID) corresponds to the minimum number of coordinates which are necessary to approximately capture the variability, revealing the structure and complexity of the manifold. This renders the ID as a geometric property \citep{valeriani2023geometry} that describes how data points are distributed within the representation space. 

Several methods have been developed to estimate intrinsic dimensionality, each differing in its underlying assumptions and formulations. \citet{levina2004maximum} uses maximum likelihood estimation to fit the likelihood on the distances from one point to each point within a \textit{fixed} neighborhood structure. If the neighborhood is set too small in a dense region, the dimensionality might be underestimated. If the neighborhood is set too large in a sparse region, it might be overestimated. \citet{farahmand2007manifold} adapts the size of the neighborhood based on the geometry of the manifold. 

\citet{facco2017estimating} exploits the expected ratio of distances between closest neighbors, observing that the distribution of distances of a point to its first neighbor is significantly smaller than to its second neighbor in lower dimensions, while in higher dimensions, the distance ratio is relatively close. By relying on the minimal information needed from the neighborhood, this approach alleviates the effects of variations in densities and curvatures within the manifold, providing stable ID estimates. 


Recent studies have investigated how intrinsic dimensionality evolves and manifests through the layers \citep{ansuini2019intrinsic}, with connections to learning dynamics \citep{aghajanyan2021intrinsic, pope2021intrinsic} and generalization \citep{birdal2021intrinsic}. \citet{ansuini2019intrinsic} demonstrated that data embedded in a high-dimensional space is progressively compressed into low-dimensional manifolds. 

\begin{table*}[!tb]
\centering
\caption{Overview of prominent techniques for differentially-private text rewriting. Scope specifies whether the method applies DP at the word-level or sentence-level. Mechanism indicates the type of privacy mechanisms. Budget refers to the recommended range of the privacy budget. Approach describes the underlying substitution mechanism, including word embeddings, causal LMs, conditional LMs, or masked LMs. Fine-tuned specifies whether the LM was explicitly fine-tuned for paraphrasing or only leveraged pre-trained representations.}

\label{tab:mechanism}
\resizebox{\linewidth}{!}{%
\begin{tblr}{
  hline{1,7} = {-}{0.08em},
  hline{2} = {2-6}{lr},
}
                                    & Scope          & Mechanism   & Budget      & Approach   & Fine-tuned \\
\citet{feyisetan2020privacy}   & Word-level     & Exponential & $\sim$ 10     & Word Embedding & no         \\
\citet{mattern2022limits}      & Sentence-level & Exponential & $\sim$ 100    & Causal LM  & yes        \\
Igamberdiev \& Habernal (2023)      & Sentence-level & Gaussian    & $\sim$ 1000 & Conditional LM & no        \\
\citet{utpala2023locally}       & Sentence-level & Exponential & $\sim$ 100    & Causal LM  & no         \\
\citet{meisenbacher2024dp} & Sentence-level & Exponential & $\sim$ 100    & Masked LM  & no         
\end{tblr}
}
\end{table*}

\section{Methodology}


We aim to investigate how privacy-preserving transformations alter the geometry of paraphrases relative to those generated without privacy guarantees.


For our experiments, we utilize \texttt{MRPC}  \citep{dolan2005automatically}, a dataset containing sentence pairs labeled for semantic equivalence. We selected sentence pairs that provide a \textit{reference} and \textit{paraphrase} to ensure a controlled basis for assessing geometric distortions in representation subspaces.

\subsection{Selection of Privacy Mechanisms}

Table \ref{tab:mechanism} outlines key characteristics of prominent approaches for differentially-private rewriting. To ensure comparability across privacy budgets, we focus on randomized mechanisms that implement the exponential mechanism. For word-level paraphrasing, we select \texttt{Madlib} \citep{feyisetan2020privacy}, which perturbs individual word in embedding space. For sentence-level paraphrasing, we select \texttt{DP-PARAPHRASE} \citep{mattern2022limits}, \texttt{DP-PROMPT} \citep{utpala2023locally}, and \texttt{DP-MLM} \citep{meisenbacher2024dp}, covering causal and masked paraphrasing with temperate sampling. \texttt{DP-PARAPHRASE} and \texttt{DP-PROMPT} are powered by fine-tuned \texttt{GPT-2} \citep{radford2019language} and pre-trained \texttt{LLaMA-3} \citep{touvron2023llama}, respectively. \texttt{DP-MLM} employs \texttt{RoBERTa} \citep{liu2019roberta}. Table \ref{tab:example} presents an example sentence from \texttt{MRPC} along with its human-authored and differentially-private paraphrases.  

\begin{table*}
\centering
\caption{Example from \texttt{MRPC} showing a sentence and its human-authored paraphrase. Note that differentially-private paraphrases at word-level are obtained using a privacy budget of $\varepsilon=25$, whereas differentially-private paraphrases at sentence-level are obtained using a privacy budget of $\varepsilon=100$.}
\label{tab:example}
\resizebox{\linewidth}{!}{%
\begin{tblr}{
  hline{1,7} = {-}{0.08em},
  hline{3} = {-}{r},
}
Sentence & Amrozi accused his brother, whom he called " the witness ", of deliberately distorting his evidence.\\
Paraphrase & Referring to him as only " the witness ", Amrozi accused his brother of deliberately distorting his evidence.\\
\citet{feyisetan2020privacy} & Amrozi accused his brother , Tyler he warn the witness  confined deliberately discolored
muse evidence.\\
\citet{mattern2022limits} & The person is Amrozi . aggression is evident even illustrates its extreme inflections over their close relative.\\
\citet{utpala2023locally} & The witness had said his wife had left him when his wife was pregnant, his second daughter was not Alis.\\
\citet{meisenbacher2024dp} & He alleged his nephew, whom he named \_ the witness " of specifically distracting his testimony.
\end{tblr}
}
\end{table*}

\subsection{Estimation of Intrinsic Dimension}


Following \citet{tulchinskii2024intrinsic}, we obtain embeddings for each word in a text using \texttt{BERT}  \citep{devlin2019bert}, treating each text as a point cloud of words spanning a manifold in the representation space. The ID of this point cloud is then estimated using \texttt{TwoNN} \citep{facco2017estimating}. To ensure that ID estimations reflect meaningful linguistic properties rather than artifacts of tokenization, we drop demarcation tokens as \texttt{<CLS>} and \texttt{<SEP>}. We also filtered short text sequences with less than 15 words and truncated long text sequences at 128 words. This stabilizes ID estimates by ensuring that estimations are based on sufficiently rich representations, while avoiding outlier effects from excessively short or long sentences.

Our investigation spans a range of privacy budgets $\varepsilon \in \{10,15,20,25,50,100\} $, allowing us to weigh the geometric distortions with respect to the desired level of privacy. Since temperature sampling is probabilistic, we repeat the paraphrasing process three times per sample at each privacy level, ensuring robust ID estimations across multiple trials and reducing variance in the distortions.


\section{Findings}

\begin{figure}[!tb]
    \includegraphics[width=0.5\textwidth]{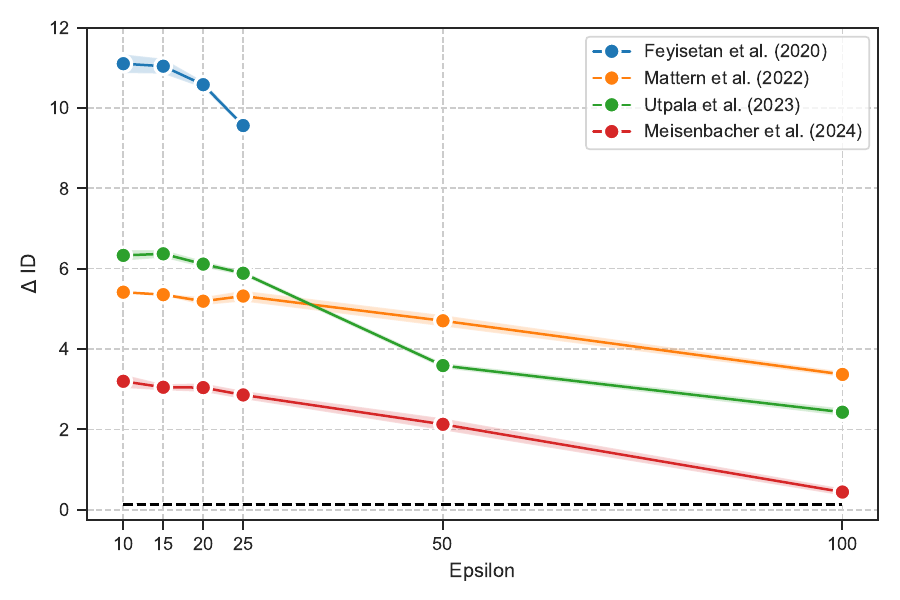}
    \caption{Shift in the estimated number of intrinsic dimensions, with a horizontal line representing a lower bound derived from human-authored paraphrases.
    }
    \label{fig:shifts}
\end{figure}

Figure \ref{fig:shifts} presents the deviation in the number of ID as a function of the privacy budget. To establish a lower bound for ID shifts, we measure the ID difference between reference sentences and their human-authored paraphrases from \texttt{MRPC}. This yields an ID shift of approximately 0.12, indicating that naturally occurring paraphrasing introduces only minimal geometric distortions in the representation space. Any privacy-preserving transformation that deviates strongly from this baseline alters the structure and complexity of text representations beyond natural variation, potentially affecting readability.


\paragraph{Word-Level Perturbation.}

Since \texttt{MADLIB} is applied at word-level, its randomized mechanism perturbs words independently, disregarding sentence structure and grammatical coherence. This results in fragmented and disorganized text, a phenomenon that can be observed through the highest ID shifts among all approaches. This observation reinforces a fundamental limitation of word-level perturbations, which induce severe distortions in representation subspaces, making them unsuitable for privacy-preserving paraphrasing.

\paragraph{Sentence-Level Perturbation.} 

Unlike \texttt{MADLIB}, which perturbs words in isolation, sentence-level perturbation incorporates context when generating paraphrases. Across all privacy budgets, sentence-level perturbation introduces significantly less distortion, as indicated by their consistently lower ID shifts. This demonstrates that leveraging LMs produces more natural paraphrases.

Among causal paraphrasing, a mixed pattern emerges depending on the privacy regime. The ID shift of \texttt{DP-PARAPHRASE} remains stable across privacy budgets, whereas \texttt{DP-PROMPT} declines more sharply. At strict privacy regimes, \texttt{DP-PARAPHRASE}, which is explicitly fine-tuned for paraphrasing, outperforms \texttt{DP-PROMPT}, which learns paraphrasing implicitly from pre-training. At more relaxed privacy regimes, however, \texttt{DP-PROMPT} surpasses \texttt{DP-PARAPHRASE} by operating more within human-like representation geometry. Since privacy is enforced via temperature sampling, this trend suggests differing sensitivity to temperature values. \texttt{DP-PARAPHRASE} handles high temperatures more effectively, whereas \texttt{DP-PROMPT} tends to generate excessively complex paraphrases. Unlike autoregressive paraphrasing, \texttt{DP-MLM} adopts masked paraphrasing, reconstructing words bidirectionally rather than generating words sequentially. \texttt{DP-MLM} clearly excels across all privacy budgets, yielding more stable representation geometry.


\paragraph{Error Propagation}

We argue that a key factor driving the divergence between causal and masked paraphrasing stems from error propagation. Causal paraphrasing perturbs text in a fixed order, where each word conditions the selection of the next word, whereas masked paraphrasing operate bidirectionally, conditioning each word substitution on both preceding and following context. When differential privacy is enforced through temperature sampling, it introduces randomness, destabilizing generation by increasing the likelihood of unnatural word choices. Once a word has been poorly substituted, the language model  must compensate to maintain fluency, leading to cascading errors which manifest in the form of drastic changes in the representation subspace. Since masked paraphrasing is not constrained by sequential consistency, distortion from a poorly chosen word does not propagate along the sentence, preventing error accumulation and producing more stable paraphrases.


\section{Conclusion}

We analyze the transformative effects of applying DP to text, focusing on how privacy constraints induce geometric distortions in the representation space. By leveraging the ID as a measure of structural complexity, we assess the extent to which prominent DP mechanisms alter latent subspaces and reshape linguistic representations. Our findings reveal that word-level DP introduces severe ID shifts, leading to drastically inflated representation manifolds. For sentence-level DP, we observe distinct differences between their representation geometry, depending on how words are substituted and whether errors from suboptimal word choices accumulate and propagate throughout a sentence.

\paragraph{Limitations.} A limitation of our inspection is that ID estimation, while a powerful tool for inspecting representation geometry of text, does not directly capture linguistic quality. Although ID shifts provide evidence of geometric distortions, connecting these distortions to measures of fluency \citep{salazar2020masked} and adequacy \citep{zhang2019bertscore, yuan2021bartscore} would complement our understanding of alterations induced by DP rewriting. 

\bibliography{custom}
\end{document}